\documentclass[review]{elsarticle}

\usepackage{subcaption}
\usepackage{lineno}
\usepackage{bbm}
\usepackage[ruled,vlined]{algorithm2e}
\usepackage{amsmath}
\DeclareMathOperator*{\argmax}{arg\,max}
\DeclareMathOperator*{\argmin}{arg\,min}
\modulolinenumbers[5]
\usepackage[table,dvipsnames]{xcolor}
\usepackage{color}
\usepackage{diagbox}
\usepackage{slashbox}
\usepackage{multirow}
\usepackage{amsfonts}
\usepackage{bigstrut}
\usepackage{subcaption}
\usepackage{soul}
\usepackage{tabularx}
\usepackage{xcolor}
\usepackage{booktabs}
\usepackage{makecell}
\usepackage{adjustbox}
\usepackage{rotating}

\colorlet{tdcolor}{yellow!35}
\sethlcolor{tdcolor}
\definecolor{bluencs}{rgb}{0.0, 0.53, 0.74}
\definecolor{bleudefrance}{rgb}{0.19, 0.55, 0.91}
\journal{Pattern Recognition}

\begin{document}
\begin{frontmatter}

\title{Counterfactual Explanation Based on Gradual Construction for Deep 
Networks} 

\author[address_1]{Hong-Gyu Jung\fnref{equal}}
\author[address_1]{Sin-Han Kang\fnref{equal}}
\author[address_2]{Hee-Dong Kim}
\author[address_3]{Dong-Ok Won}
\author[address_2]{Seong-Whan Lee\corref{mycorrespondingauthor}}
\fntext[equal]{Equal contribution}
\cortext[mycorrespondingauthor]{Corresponding author}
\ead{sw.lee@korea.ac.kr}

\address[address_1]{Department of Brain and Cognitive Engineering, Korea University, Anam-dong, Seongbuk-gu, Seoul, 02841, Korea}
\address[address_2]{Department of Artificial Intelligence, Korea University, Anam-dong, \\ Seongbuk-gu, Seoul, 02841, Korea}
\address[address_3]{Department of Artificial Intelligence Convergence, Hallym University, Chuncheon, Gangwon, 24252, Korea}

\begin{abstract}
To understand the black-box characteristics of deep networks, counterfactual explanation that deduces not only the important features of an input space but also how those features should be modified to classify input as a target class has gained an increasing interest. The patterns that deep networks have learned from a training dataset can be grasped by observing the feature variation among various classes. However, current approaches perform the feature modification to increase the classification probability for the target class irrespective of the internal characteristics of deep networks. This often leads to unclear explanations that deviate from real-world data distributions. To address this problem, we propose a counterfactual explanation method that exploits the statistics learned from a training dataset. Especially, we gradually construct an explanation by iterating over masking and composition steps. The masking step aims to select an important feature from the input data to be classified as a target class. Meanwhile, the composition step aims to optimize the previously selected feature by ensuring that its output score is close to the logit space of the training data that are classified as the target class. Experimental results show that our method produces human-friendly interpretations on various classification datasets and verify that such interpretations can be achieved with fewer feature modification.
\end{abstract}


\end{frontmatter}

\section{Introduction}

Although deep networks exhibit remarkable performances in various tasks, the internal complexity of the models results in a transparency issue. Specifically, considering that deep networks comprise various non-linear functions, the internal mechanisms for the networks to produce an output are difficult to analyze and using such models in high risk tasks, such as credit evaluation \cite{rosenberg1994quantitative, hsu2019enhanced} and autonomous driving \cite{xu2017end, garg2016unsupervised} that require significant reliability and stability, is challenging owing to their lack of interpretability. In addition, the EU general data protection regulation \cite{wachter2018counterfactual} officially requires that the decision of a deep network can be explained. To comply with these technical and legal requirements, recent studies have developed algorithms that provide explanations for the decisions of deep networks.

Many of those exploit feature attribution methods \cite{bach2015pixel, fong2017interpretable, dabkowski2017real, nam2020relative}, which visualize the important features of an input data that lead the model to make its prediction. However, feature attribution methods are only concerned in interpreting the given input data and the predicted class; thus, the discriminative features that the model has learned to distinguish different classes cannot be directly interpreted.

To address this problem, we focus on generating counterfactual explanations. Given an input data that are classified as a class from a deep network, our goal is to perturb the subset of features in the input data such that the model is forced to predict the perturbed data as a target class. Hence, we can identify crucial features required for pre-trained networks to classify input into the target class. Fig. 1 demonstrates the difference between the feature attribution explanation and counterfactual explanation. Given the digit image with the class of ``7'', layer-wise relevant propagation (LRP) \cite{bach2015pixel}, one of the representative feature attribution methods\footnote{A feature attribution method is also referred to as a factual explanation method.}, highlights the important pixels to classify it as ``7'' by coloring them with red. Meanwhile, our counterfactual explanation method generates a perturbed image that is classified as a target class.
\begin{figure}[t]

    \centering
    \begin{subfigure}{.35\textwidth}
            \includegraphics[width=1\linewidth]{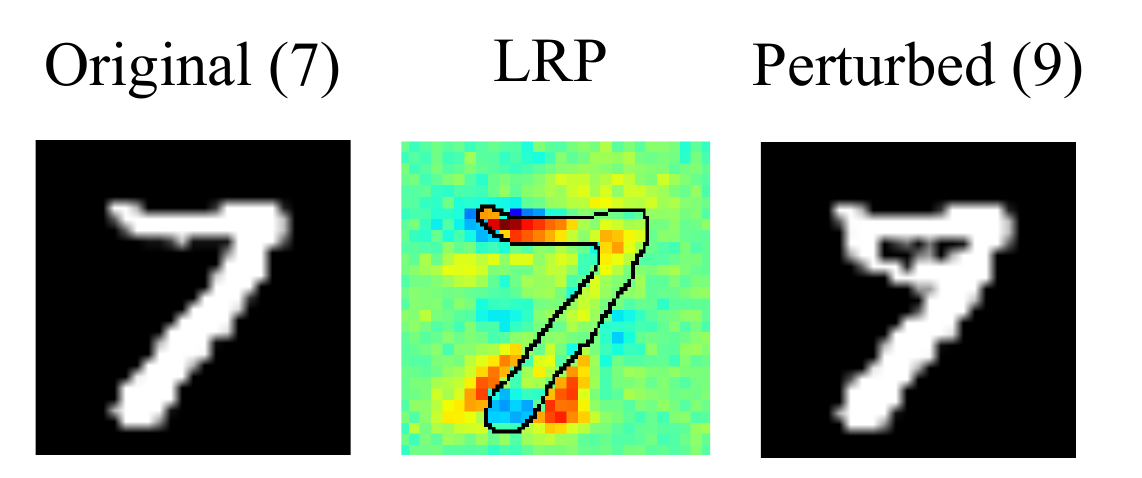}  
    \end{subfigure}
    \begin{subfigure}{.6\textwidth}
        \begin{adjustbox}{width=1\textwidth}
        \begin{tabular}{ccc}\hline
            \scriptsize{ }&  \scriptsize{Text} & \scriptsize{Prediction}\\ \hline
            \scriptsize{Original} & \scriptsize{Perfect film from beginning to end} & \scriptsize{\textcolor{red}{Positive}}\\
             \scriptsize{Perturbed} &\scriptsize{\textcolor{blue}{Shoddy} film from beginning to end} & \scriptsize{\textcolor{blue}{Negative}}\\ \hline
        \end{tabular}
        \end{adjustbox}
    \end{subfigure}
    \caption{(a) Comparison between counterfactual explanation and layer-wise relevant propagation (LRP) \cite{bach2015pixel} that is a feature attribution method. The classification results from a pre-trained network are presented above the images. (b) Example of counterfactual explanation using the IMDB sentiment analysis dataset \cite{maas2011learning} with the prediction results.}
    \label{fig:Introduction}
\end{figure}
Thus, we can have an in-depth understanding of a network as the method verifies what a network has learned to differentiate between ``7'' and ``9'' \footnote{Note that a target class is selected according to our intention to analyze the model \cite{Lecun98gradient-basedlearning}.}. In addition to the technical context, counterfactual explanation can be effectively used in real-world applications. For example, if a credit company using an AI system refuses a loan, our method can provide people with factors (e.g. loan amount and credit score) that are important to the assessment and which factors should be modified to meet some threshold values for loan approval. This aspect will be discussed in the experimental section.

Meanwhile, the perturbed data for counterfactual explanation should have two desirable properties. (i) \textbf{Explainability}: a generated explanation should be naturally understood by humans. Considering the example in Fig.~\ref{fig:Introduction}(b), a pre-trained model predicts the original text as the positive class, whereas our counterfactual explanation method converts the word ``Perfect'' into ``Shoddy'' to classify the text as the negative class. Clearly, we observe that the trained model regards ``Perfect'' and ``Shoddy'' as crucial features for the prediction of positive and negative classes. (ii) \textbf{Minimality}: only a few features should be perturbed. If we generate entirely different features from the original data to alter the original classification, the relation with the original data cannot be determined and this is, therefore, only regarded as generation but not explanation. As an example, assume that the perturbed text “\textit{Shoddy film from beginning to end}” is changed to “\textit{The film is shoddy}” by a counterfactual method. In this case, the changed text can produce an alternative decision but the discriminative features learned by the deep network are difficult to identify. 

Although several works for counterfactual explanation have been proposed, they have limitations to employing their methods in various applications. Specifically, reference-based feature generation approaches \cite{chang2018explaining, goyal2019counterfactual} were developed but the methods can be applied only to the image domain. Though domain-agnostic counterfactual explanations may be available \cite{wachter2018counterfactual, dhurandhar2018explanations}, they fail to satisfy the two properties and tend to provide unclear explanations that can be regarded as adversarial attack. In other words, the generated explanations appear similar to ~original data but are predicted as target classes. In the following sections, we verify that such a phenomenon can be resolved by considering the logit distribution of training data when generating explanation.

Herein, we propose a counterfactual explanation method based on gradual construction that considers the statistics learned from training data. We particularly generate counterfactual explanation by iterating over masking and composition steps. Given an input data, the masking step aims to select the most effective subsets of features to classify the input into a target class. To achieve this, we calculate the directive derivative with respect to the input data and choose the features that have higher sensitivity. Then, the composition step optimizes the value of the selected subsets of features by ensuring that the logit score is close to the logit distribution of the training data that are classified as the target class. This prevents the optimized features from being generated in an unpredictable distribution.

We conduct extensive experiments on text, image and finance datasets such as IMDB sentiment analysis \cite{maas2011learning}, MNIST \cite{lecun1998gradient}, HELOC \cite{FICO} and UCI Credit Card \cite{yeh2009comparisons} datasets. Experimental results show that our counterfactual method produces human-friendly explanation using much fewer input features compared to state-of-the-art methods.

\section{Related work} \label{sect:relatedworks}

Many explanation methods have been developed to produce an intuitive visualization map on a given input data. Gradient-based explanation methods \cite{bach2015pixel, nam2020relative, shrikumar2017learning,  sundararajan2017axiomatic, kauffmann2020towards, montavon2017explaining} extract a representative value for each pixel by exploiting a backward operation in neural networks. Activation-based methods \cite{zhou2016learning, selvaraju2017grad, chattopadhay2018grad, liu2020towards} utilize activation maps of the convolution layer in CNNs to provide visual explanations. Reference-based explanation methods \cite{ fong2017interpretable, dabkowski2017real, ribeiro2016should, fong2019understanding, wagner2019interpretable} compute the sensitivity of prediction scores with respect to perturbed data that are generated from masking the subset features of the original input data and replacing them with reference values such as blurred pixels, mean pixels and random noise. Furthermore, decomposition-based methods \cite{zhou2018interpretable, chen2019looks} decompose an activation map for a classification into multiple components. Each component highlights segmented regions in the input image and has the associated importance score for the classification. In summary, all these approaches aim to generate a visualization map that highlights important regions of the input data that have impact on the prediction of deep networks. Meanwhile, we focus on creating a counterfactual explanation that indicates which features in the input data should be modified and how to generate a target class.

C. H. Chang et al. \cite{chang2018explaining} recently proposed a counterfactual explanation method by masking and replacing certain image regions with artificially generated data such as blurred or generative adversarial network (GAN)-based images. Y. Goyal et al. \cite{goyal2019counterfactual} allowed a user to manually select a reference image whose prediction is a target class. Then, they aim to replace some regions of the original image with certain regions of the reference image to generate a counterfactual explanation. ABELE \cite{guidotti2019black} exploits a genetic algorithm to generate several samples around an input latent vector using an auto-encoder, and employs these samples to learn a decision tree. After that, it finds a latent vector that changes from an original class to another class and decodes the vector to generate counterfactual explanation. However, these previous studies can be applied only to the image domain. 

Meanwhile, there exist counterfactual explanation methods \cite{guidotti2019factual, laugel2018comparison} for a binary classifier. LORE \cite{guidotti2019factual} is a preceding work of ABELE \cite{guidotti2019black} and generates several samples around an input without the auto-encoder to train a decision tree. By using the path of the tree, both factual and counterfactual explanations can be found. Growing sphere \cite{laugel2018comparison} searches for a counterfactual explanation by using the L2 distance to generate samples that gradually move away from the input and checking whether the  classification result changes.

Although domain-agnostic counterfactual explanations  \cite{wachter2018counterfactual, dhurandhar2018explanations} have been developed to handle multi-class classification in addition to being domain agnostic, the methods tend to generate adversarial data rather than the one to interpret model characteristics as shown in Fig.~\ref{fig:MNISTResult_Comp}(b). As we will present in the following sections, our method overcomes this phenomenon by generating perturbed images whose logit scores follows the logit distribution of a training dataset. 

\section{Methods}
\subsection{Problem definition}

\begin{figure}[t]
\begin{center}
\includegraphics[width=1\linewidth]{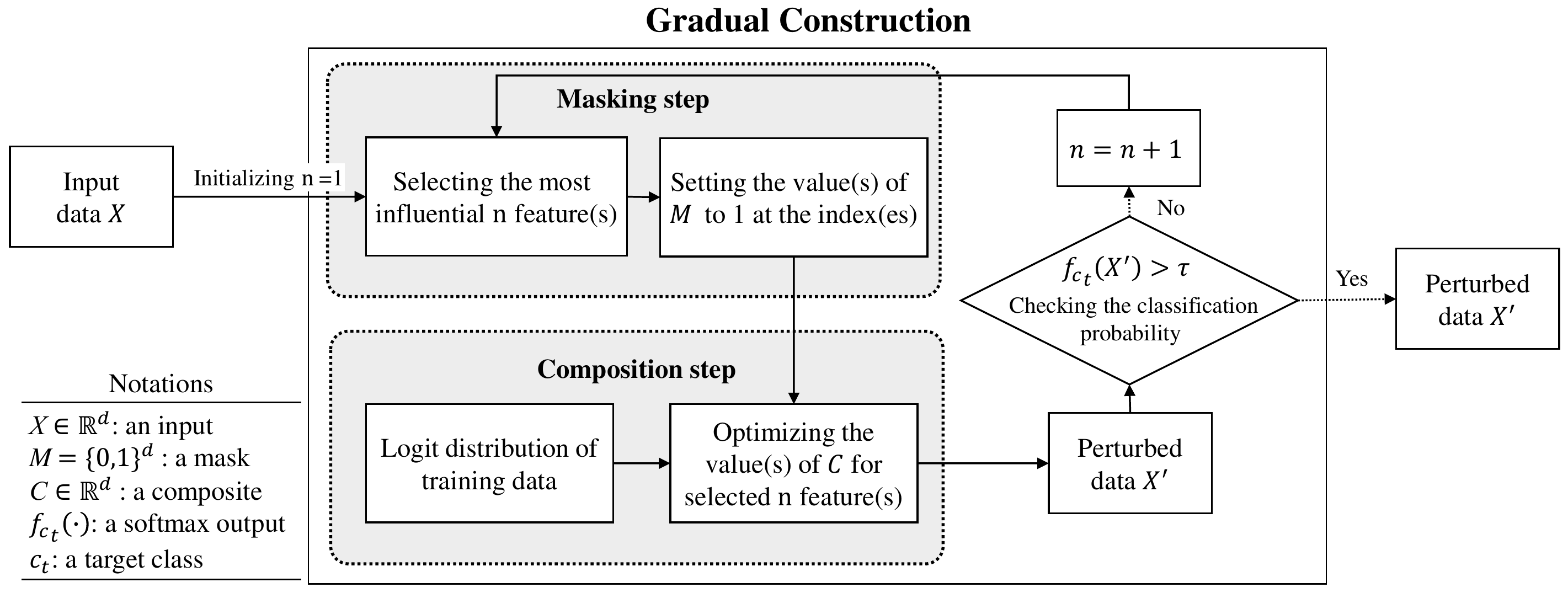}
   \caption{Overall procedure of the proposed counterfactual explanation. Our method builds gradual construction, iterating over masking steps and composite steps alternatively. These procedures are repeated until the classification probability of the perturbed data is over $\tau$ for a target class $c_t$.}
	\label{fig:OverallFigure}
   \end{center}
\end{figure}

In this section, we outline a domain-agnostic method through gradual construction of counterfactual explanations. Given an input data $X \in \mathbb{R}^{d}$ that is classified as a class $c_o$ under a pre-trained deep network $f$, we aim to perturb only the minimal subset features of $X$ to change its decision into a target class $c_t$. Specifically, in order to generate a perturbed data $X'$, we define a binary mask $M$ and a composite $C \in \mathbb{R}^{d}$. The binary mask $M=\{0,1\}^{d}$ indicates whether to replace subset features of $X$ with the composite $C$ or to preserve the features of $X$. The composite $C$ represents newly generated feature values that will be replaced into a perturbed data $X'$ instead of the original input features. Thus, we can formalize the perturbed data $X'$ as the mask $M$ and the composite $C$ as follows:
\begin{equation}
X'=(1-M)\circ X+M \circ C,
\end{equation}
where $M$ is initialized as a vector of all zeros and $\circ$ denotes the element-wise multiplication. To produce a perturbed data $X'$ whose prediction will be a target class $c_t$, we progressively search for an optimal mask and a composite. To this end, our method builds gradual construction that iterates over the masking and composition steps until the desired classification score $\tau$ for the target class $c_t$ is obtained. The goal of the masking step is to select an important feature to change the original decision into the target class $c_t$. After selecting the important feature, a value of mask $M$ corresponding to the position of the feature is changed from 0 to 1. Then, the composition step optimizes a value of $C$ for the selected feature in order to improve the output score of a target class and produce more interpretable explanations. In the following, we formally express both masking and composition steps and present our algorithm. Fig. \ref{fig:OverallFigure} shows the overall procedure of our method.

\subsection{Masking step}

The goal of the masking step is to select the most influential feature to produce a target class from a pre-trained network as follows:
\begin{equation} \label{eq:MaskingStep_1}
i^{*}=\argmax_{i} f_{c_t}(X + \delta e_i),
\end{equation} 
where $e_i$ is a one-hot vector whose value is $1$ only for the $i$-th element, $\delta$ is a non-zero real value and $f_{c_t}$ denotes the classification score for the target class $c_t$.
We first suppose $\delta =\bar{\delta}h$ where $h$ is a non-zero and infinitesimal value and $\bar{\delta}$ is a proper scalar to match the equality. Then, the objective function is approximated as the directional derivative with respect to $X$.
\begin{equation}\label{eq:MaskingStep_2}
\begin{aligned}
f_{c_t}(X + \delta e_i)
&=f_{c_t}(X + \delta e_i)-f_{c_t}(X)+f_{c_t}(X)\\
&=f_{c_t}(X + \bar{ \delta} h e_i)-f_{c_t}(X)+f_{c_t}(X)\\
&=\frac{f_{c_t}(X + \bar{\delta} e_i h )-f_{c_t}(X)}{h} h+f_{c_t}(X)\\
&\approx \bigtriangledown f_{c_t}(X) \bar{\delta} e_i h+f_{c_t}(X)\\
&=\bigtriangledown f_{c_t}(X) \delta e_i+R.
\end{aligned}
\end{equation}
Note that $f_{c_t}(X)$ that is not relevant to $i$ is regarded as a constant $R$. Since the $\delta$ is a real value, we separately consider positive and negative cases in order to find an optimal $i^{*}$. 
\begin{equation}\label{eq:MaskingStep_3}
\begin{aligned}
i^{*}=\left\{\begin{matrix}
 max( \bigtriangledown f_{c_t}(X))_{i} , & if \; \, \delta>0 \\ 
 min( \bigtriangledown f_{c_t}(X))_{i} , & otherwise.
\end{matrix}\right.
\end{aligned}
\end{equation}
Here, the $max(\cdot)_{i}$ function returns an index that has a maximum value in the input vector and $min(\cdot)_{i}$ is similarly defined. 

However, given that the $\delta$ value is determined at the composition step, it is not known which function should be used between the maximum and minimum operators to select the optimal $i^*$. Thus, we choose a sub-optimal index as
\begin{equation}\label{eq:MaskingStep_4}
\begin{aligned}
\hat{i}^{*}=max(\lvert \bigtriangledown f_{c_t}(X) \rvert)_{i}.
\end{aligned}
\end{equation}
Although the sub-optimal choice is possible to induce more iterations for gradual construction, experimental results show that our method can efficiently produce counterfactual explanations with fewer features than state-of-the-art methods. In summary, each masking step selects an index in the descending order by calculating Eq. \ref{eq:MaskingStep_4} and changes the zero value of mask $M$ into one.


\subsection{Composition step}
After selecting the input feature to be modified, the composition step optimizes the feature value to ensure that the deep network classifies the perturbed data $X'$ as the target class $c_t$. To achieve this, previous works \cite{wachter2018counterfactual, dhurandhar2018explanations} have proposed an objection function to improve the output score of $c_t$ as follows:
\begin{equation}\label{eq:PriorLoss}
\argmax_{\epsilon} f_{c_{t}}(X+\epsilon)+R_\epsilon,
\end{equation}
where $\epsilon=\{\epsilon_1, ..., \epsilon_d\}$ is a perturbation variable and $R_\epsilon$ is a regularization term.


\begin{figure}[t]
\begin{center}
\includegraphics[width=1\linewidth]{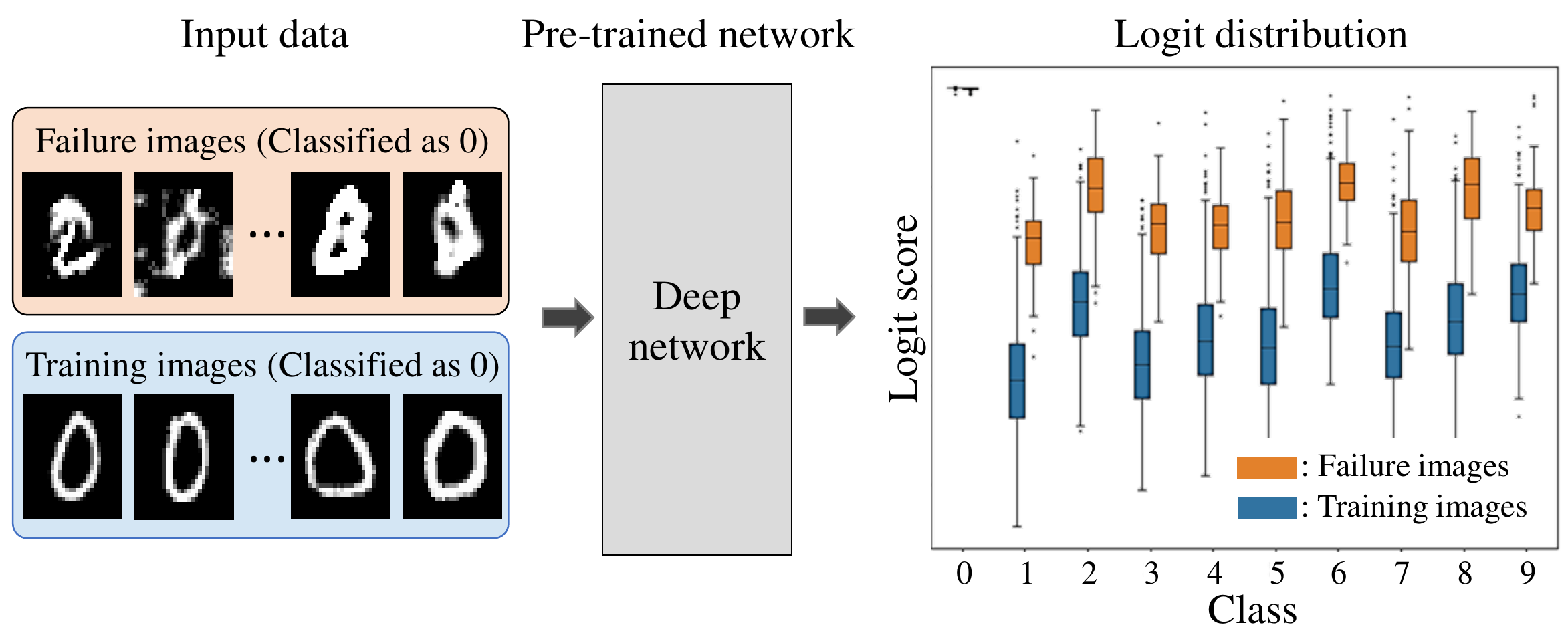}
   \caption{Comparison of logit distributions between 100 real training images and 100 failure images of CEM \cite{dhurandhar2018explanations} that are classified as the class ``0'' under a trained model. The blue box plot is for the training data and the orange box plot represents the images of failure cases.}
	\label{fig:LogitDistribution}
   \end{center} 
\end{figure}

\begin{algorithm}[!t]
\SetAlgoLined
\nl \KwIn{ 
  \, \, \textbullet~$X\in\mathbb{R}^{d}$: an input data

  \, \, \textbullet~$c_t$: a target class
  
  \, \,  \textbullet \, $\tau$: the desired classification probability for the target class
  
  \, \,  \textbullet \, $\sigma$: the number of iteration

}

\nl \bf Initialization:

  \, \, \textbullet \, {\normalfont Mask} $M\in \mathbb{R}^{d}$ {\normalfont and} $M_i=0 \; \, \forall i$ 
  
  \, \, \textbullet \, {\normalfont Composite} $C\in  \mathbb{R}^{d}$ {\normalfont and} $C_j \sim  N(0,1) \; \, \forall j$
  
  \, \, \textbullet \, {\normalfont Perturbed data } $X'=(1-M)\circ X+M\circ C$
  
  \, \, \textbullet \, {\normalfont The number of perturbed features } $n=1$
\texttt{\\}
\nl \bf {While $f_{c_{t}}(X') < \tau $:}

	\bf \ \ \ 1) Masking step
	
	\nl \ \ \ \ \ \  $i^{*} \gets$ {\normalfont an index of the} $n$ {\normalfont highest value in} $|\bigtriangledown f_{c_t}(X)|$
    
    \nl \ \ \ \ \ \ $M_{i^{*}} \gets 1$

    \bf \ \ \ 2) Composition step
    
    \nl \ \ \ \bf \ \ \ for $m=1$ to $\sigma$ do
    
    \nl \ \ \ \ \ \ \ \ \ $C \gets \argmin\limits_{C} \left\Vert \sum_{k=1}^{K} \left( f'_{k}(X')-\frac{1}{N}\sum_{i=1}^{N} f'_{k}(X_{i,c_t}) \right) \right\Vert_2$
    
    \ \ \ \ \ \ \ \ \ \qquad\qquad\qquad $+ \lambda \left\Vert X' - X \right\Vert_2$

    \nl \ \ \ \bf \ \ \ $n \gets n+1$
    
    \caption{{\bf Gradual construction} \label{pseudocode}}
\nl \KwOut{$X'$}
\end{algorithm}
Although it is possible for the objective function in Eq.~\ref{eq:PriorLoss} to generate interpretable counterfactual explanations, we find that it also causes an adversarial attack as shown in Fig.~\ref{fig:MNISTResult_Comp}(b). To analyze the reason, we accumulated numerous failure cases of CEM \cite{dhurandhar2018explanations} for the MNIST dataset. Then, we compared the distributions of logit scores (before the softmax layer) for each failure case and the training images that are classified as $c_t$ from a pre-trained network. As shown in Fig.~\ref{fig:LogitDistribution}, we discovered that there exist a notable difference between the two distributions. When we examine the logit score distributions of the training images and failure images that are classified as a target class 0, we can observe that the logit value of failure cases from classes 1 to 9 are generally higher than the training data. Thus, we regard failure cases as the result of an inappropriate objective function that maps the perturbed data onto a different logit space from the training data. To solve this problem, we instead force the logit space of $X'$ to belong to the space of training data as follows:
\begin{equation} \label{eq:ProposedLoss}
\argmin_{C} \left\Vert \sum_{k=1}^{K} \left( f'_{k}(X')-\frac{1}{N}\sum_{i=1}^{N} f'_{k}(X_{i,c_t}) \right) \right\Vert_2+\lambda \left\Vert X' - X \right\Vert_2,
\end{equation}
where $X'=(1-M)\circ X+M\circ C$, $K$ is the number of classes, $f'_k$ represents a logit score for a class $k$, $X_{i, c_t}$ denotes $i$-th training data that is classified into a target class $c_t$. $N$ denotes the number of randomly sampled training data. In addition, to prevent from generating a totally different data from an input, we add a regularizer $\lambda$ to encourage the values of $X'$ to be close to the input data $X$. As a result, Eq.~\ref{eq:ProposedLoss} makes the composite $C$ to improve the probability of $c_t$ and also pushes the perturbed data towards belonging to the logit score distribution of a training data.


Overall, gradual construction iterates over the masking and composition steps until the classification probability of a target class is reached to a hyperparameter $\tau$. We present a pseudo-code in Algorithm \ref{pseudocode}.

\section{Experiments} \label{sect: exp}

\begin{table}[!t]
\centering
\caption{The statistics of the IMDB, MNIST, HELOC and UCI Credit Card datasets.}
\begin{adjustbox}{width=1\textwidth}
\renewcommand{\arraystretch}{1.0}
\label{table_statistics}
\begin{tabular}{c|cccc}
\hline
Dataset & Type & \# of training data & \# of test data & \# of classes \bigstrut\\
\hline
IMDB \cite{maas2011learning} & Text & 25,000 & 25,000 & 2 (Positive/Negative) \bigstrut[t]\\
MNIST \cite{lecun1998gradient} & Image & 60,000 & 10,000 & 10 \\
HELOC \cite{FICO} & Tabular & 7,402 & 2,468 & 2 (Approval/Refual) \\
UCI Credit Card \cite{yeh2009comparisons} & Tabular & 22,500 & 7,500 & 2 (Approval/Refual) \bigstrut[b]\\
\hline
\end{tabular}%
\end{adjustbox}
\end{table}

We provide extensive experiments on text, image and finance datasets as follows. (1) Comparison with a feature attribution method on the text domain with the IMDB sentiment analysis dataset \cite{maas2011learning}. (2) Comparison with counterfactual explanation methods on the image domain with the MNIST dataset \cite{lecun1998gradient} and (3) on the finance domain with the HELOC \cite{FICO} and UCI Credit Card \cite{yeh2009comparisons} datasets. (4) Ablation study on the image and finance datasets to validate the effectiveness of our loss function that prevents a generated explanation from being adversarial data. The statistic of each dataset is provided in Table \ref{table_statistics}.


\subsection{Experimental setting}
We trained a multi-layer perceptron (MLP) or a convolutional neural network (CNN) that is specified below for each dataset. To generate a composite, we used the Adam optimizer \cite{kingma2014adam} and set the learning rate to 0.1. Training iteration was set to $1,000$ and $500$ for the MNIST dataset and the other datasets, respectively. We used the hyper-parameters $N$=100 and $\lambda$=0.3.

\subsection{IMDB}
The IMDB dataset \cite{maas2011learning} is a movie reviews dataset for sentiment classification. As we need to rely on word-to-embedding pairs for given texts, new word-to-embedding pairs for counterfactual explanation are generated to change the classification result to the alternative class. 

\subsubsection{Pre-trained network}
Given that the dataset is composed of words, word embeddings created by  GloVe \cite{pennington2014glove} were used as input. Then, we trained a CNN model with three convolution layers, three max-pooling layers, a Dropout layer \cite{JMLR:v15:srivastava14a} and a fully-connected layer. The minimum word count in a movie review is restricted to five for the input. When its word number is lower than 5, the word `pad' is added to match the requirement. In particular, this CNN model achieved a 85.4\% test accuracy. 
\subsubsection{Algorithmic details}
As we used GloVe \cite{pennington2014glove}, unique word-to-embedding pairs exist. However, after applying our explanation method, such embeddings are perturbed and thus, do not match the GloVe word-embedding pairs. Thus, we calculated the distance between the perturbed embedding and the GloVe embeddings. Finally, the word with the minimum distance was produced from the unique pair for explanation. As a hyper-parameter in Algorithm \ref{pseudocode}, the target probability $\tau$ is set to 0.9.

\begin{table}[t]
\begin{adjustbox}{width=1\textwidth}
\begin{tabular}{ccc}
\hline
        \textbf{Method}        & \textbf{Text data} &  \textbf{Prediction} \\ \hline 
LRP &  \makecell{It was one of \textcolor{RedOrange}{the} \textcolor{red}{best} \textcolor{RedOrange}{theatre} \textcolor{Melon}{experiences I} have ever had}
        &  \textcolor{red}{Positive}\\ 

Ours  &    \makecell{It was one of the \textcolor{blue}{dreadful} theatre experiences I have ever had}
       &  \textcolor{blue}{Negative} \\ \hline

LRP &  \makecell{This film \textcolor{RedOrange}{is} \textcolor{red}{great} \textcolor{RedOrange}{and} \textcolor{Melon}{great}}         & \textcolor{red}{Positive} \\ 

 Ours & \makecell{This film is \textcolor{blue}{terrible} and great}          & \textcolor{blue}{Negative} \\  \hline
LRP &   \makecell{\textcolor{bluencs}{The film} \textcolor{bleudefrance}{is} \textcolor{blue}{awful}}
       & \textcolor{blue}{Negative} \\ 
Ours  &    \makecell{The film is \textcolor{red}{truly}}      & \textcolor{red}{Positive} \\   \hline
\end{tabular}
\end{adjustbox}
\caption{Comparison on the explanations of factual and counterfactual methods using the IMDB dataset \cite{maas2011learning}. LRP \cite{bach2015pixel} that is a factual explanation method highlights important words to classify the texts. The color opacity in LRP represents the importance of the words and red and blue colors indicate positive and negative degrees. On the contrary, our method generates a counterfactual explanation by choosing the most important word and changing the word to the another to produce the alternative prediction. The changed word in our results is represented as a blue or red color.}
\label{fig:IMDBResult_Comp}
\end{table}

\begin{table}[t]
\begin{adjustbox}{width=1\textwidth}
\renewcommand{\arraystretch}{1}
\Huge
\begin{tabular}{ccc}
\hline
 \multicolumn{2}{c}{\textbf{Text data}} &  \textbf{Prediction} \\ \hline 
 Original text &  \makecell{The ultimate homage to a great film actress. The film is a masterpiece of poetry on the screen. Like \\ great poetry it is \textbf{\textcolor{red}{timeless}}.  Direction, cast, screenplay, music, lyrics, in fact all the norms for movie… }
        &  \textcolor{red}{Positive}\\ 

 Ours (100-dim)  &    \makecell{The ultimate homage to a great film actress. The film is a masterpiece of poetry on the screen. Like \\ great poetry it is \textbf{\textcolor{blue}{nauseating}}. Direction, cast, screenplay, music, lyrics, in fact all the norms for movie…}
       &  \textcolor{blue}{Negative} \\ 
       
Ours (300-dim) &    \makecell{The ultimate homage to a great film actress. The film is a masterpiece of poetry on the screen. Like \\ great poetry it is \textbf{\textcolor{blue}{trite}}. Direction, cast, screenplay, music, lyrics, in fact all the norms for movie…}
       &  \textcolor{blue}{Negative} \\ \hline

 Original text &  \makecell{I was pretty disappointed in what I believe was one of Audrey Hepburn's last movies. I 'll always \\ love John Ritter best in slapstick. He was just too \textbf{\textcolor{blue}{pathetic}} here...
}         & \textcolor{blue}{Negative} \\ 

  Ours (100-dim) & \makecell{I was pretty disappointed in what I believe was one of Audrey Hepburn's last movies. I 'll always \\ love John Ritter best in slapstick. He was just too \textbf{\textcolor{red}{delightful}} here...}          & \textcolor{red}{Positive} \\
  
Ours (300-dim) & \makecell{I was pretty disappointed in what I believe was one of Audrey Hepburn's last movies. I 'll always \\ love John Ritter best in slapstick. He was just too \textbf{\textcolor{red}{wonderful}} here...}          & \textcolor{red}{Positive} \\  \hline

 Original text &   \makecell{The mood of the film is captured perfectly by the camera-work and ( lack of ) lighting. A \textbf{\textcolor{red}{great}} discourse...}
       & \textcolor{red}{Positive} \\ 
 Ours (100-dim)  &    \makecell{The mood of the film is captured perfectly by the camera-work and ( lack of ) lighting. A \textbf{\textcolor{blue}{dreadful}} \\ discourse...
}      & \textcolor{blue}{Negative} \\   

Ours (300-dim)  &    \makecell{The mood of the film is captured perfectly by the camera-work and ( lack of ) lighting. A \textbf{\textcolor{blue}{bad}} \\ discourse...
}      & \textcolor{blue}{Negative} \\   

\hline
 Original text &  \makecell{Yes, The Southern Star features a pretty \textbf{\textcolor{blue}{forgettable}} title tune sung by that heavy set crooner Matt \\ Monro. It pretty much establishes the tone for this bloated and rather dull feature, ...
}         & \textcolor{blue}{Negative}  \\   

Ours (100-dim)  & \makecell{Yes, The Southern Star features a pretty \textbf{\textcolor{red}{memorable}} title tune sung by that heavy set crooner Matt \\  Monro . It pretty much establishes the tone for this bloated and rather dull feature, ... }          &  \textcolor{red}{Positive}\\

Ours (300-dim)  & \makecell{Yes, The Southern Star features a pretty \textbf{\textcolor{red}{memorable}} title tune sung by that heavy set crooner Matt \\  Monro . It pretty much establishes the tone for this bloated and rather dull feature, ... }          &  \textcolor{red}{Positive}\\
\hline
\end{tabular}
\end{adjustbox}
\caption{Counterfactual explanations on the texts of the IMDB test dataset \cite{maas2011learning}. The changed word in our results is represented as a blue or red color.}
\label{IMDBResult_More_Qualitative}
\end{table}

\subsubsection{Results}
To compare our method with LRP \cite{bach2015pixel} that is one of the representative feature attribution methods, we generated several text data for test as shown in Table~\ref{fig:IMDBResult_Comp}. LRP highlights the important words of the original text which leads the model to make its prediction. To be specific, LRP in the first column of Table~\ref{fig:IMDBResult_Comp} explains that the subset words ``the best theatre experiences I'' in the original text contribute to produce the positive class and the most important word among them is `best'. However, it is not clear how many words a pre-trained network needs to keep its prediction and it is not possible to explicitly know the meaning of the color opacity among the highlighted words. In contrast, our method changes `best' into `dreadful'. This result implies that the word `best' is critical to classify the text to positive regardless of the other words, and the changed word `dreadful' mainly contributes to produce an alternative class. Thus, we argue that our method can provide not only the critical regions of an input data in order to output its prediction but also what feature(s) should be changed to produce an alternative decision in terms of interpretability.

The another interesting point is illustrated in the second column of Table~\ref{fig:IMDBResult_Comp}. Our method reveals that the pre-trained model thinks only the first word `great' as an important element to its prediction, so that substituting `great' into `terrible' forces the model to classify the text as the negative class. Meanwhile, the last column of Table~\ref{fig:IMDBResult_Comp} shows the weakness of our method in the text data. The perturbed word `truly' can be considered positive by humans, but this word is grammatically incorrect. In other words, our method does not consider a grammatical error when generating counterfactual explanations.

The more qualitative results on the texts provided by the IMDB dataset are presented in Table \ref{IMDBResult_More_Qualitative}. The original data is randomly selected from the test set and the highlighted part indicates the word before and after perturbation using our method. As an ablation study, we also provide the results with two different lengths of GloVe \cite{pennington2014glove} embeddings. The average numbers of changed words when using each embedding in the test set were $1.94\pm5.67$ and $1.52\pm0.94$, respectively. As indicated in the standard deviation, our method operates more stably with the 300-dimensional embedding due to its higher discriminative power.

\begin{figure*}[t]
\begin{center}
\includegraphics[width=\linewidth]{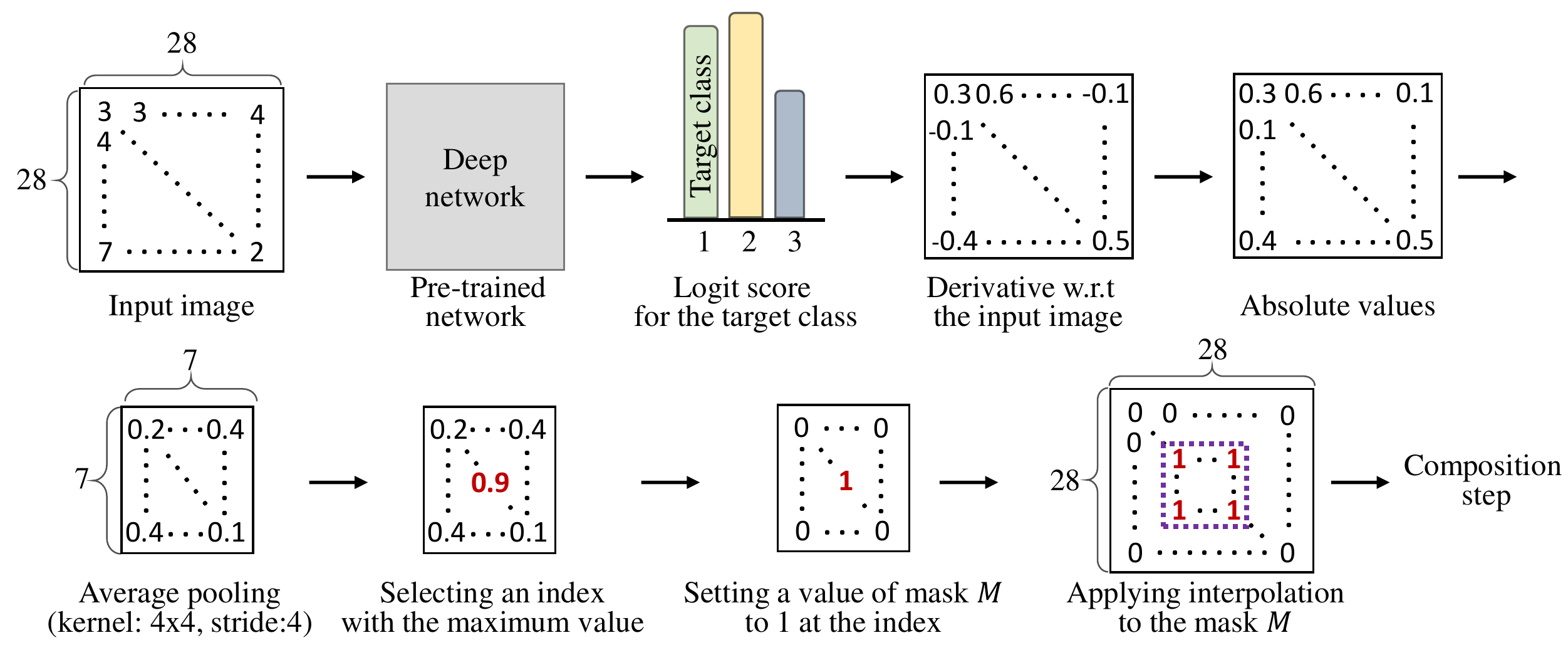}
   \caption{Detailed procedure of the masking step in the MNIST dataset \cite{lecun1998gradient}.}
	\label{fig:Making_MNIST}
   \end{center}
\end{figure*}

\subsection{MNIST}
The MNIST dataset \cite{lecun1998gradient} is composed of hand-written digits from 0 to 9. Counterfactual explanation is provided by generating new pixels that change the classification result to a target class. 

\subsubsection{Pre-trained network}
We trained a simple CNN for digit classification, which consists of {two sets of convolution-convolution-pooling layers followed by three fully-connected layers.} The CNN obtained a 98.4\% test accuracy.
\begin{figure*}[t]
\begin{center}
\includegraphics[width=1\linewidth]{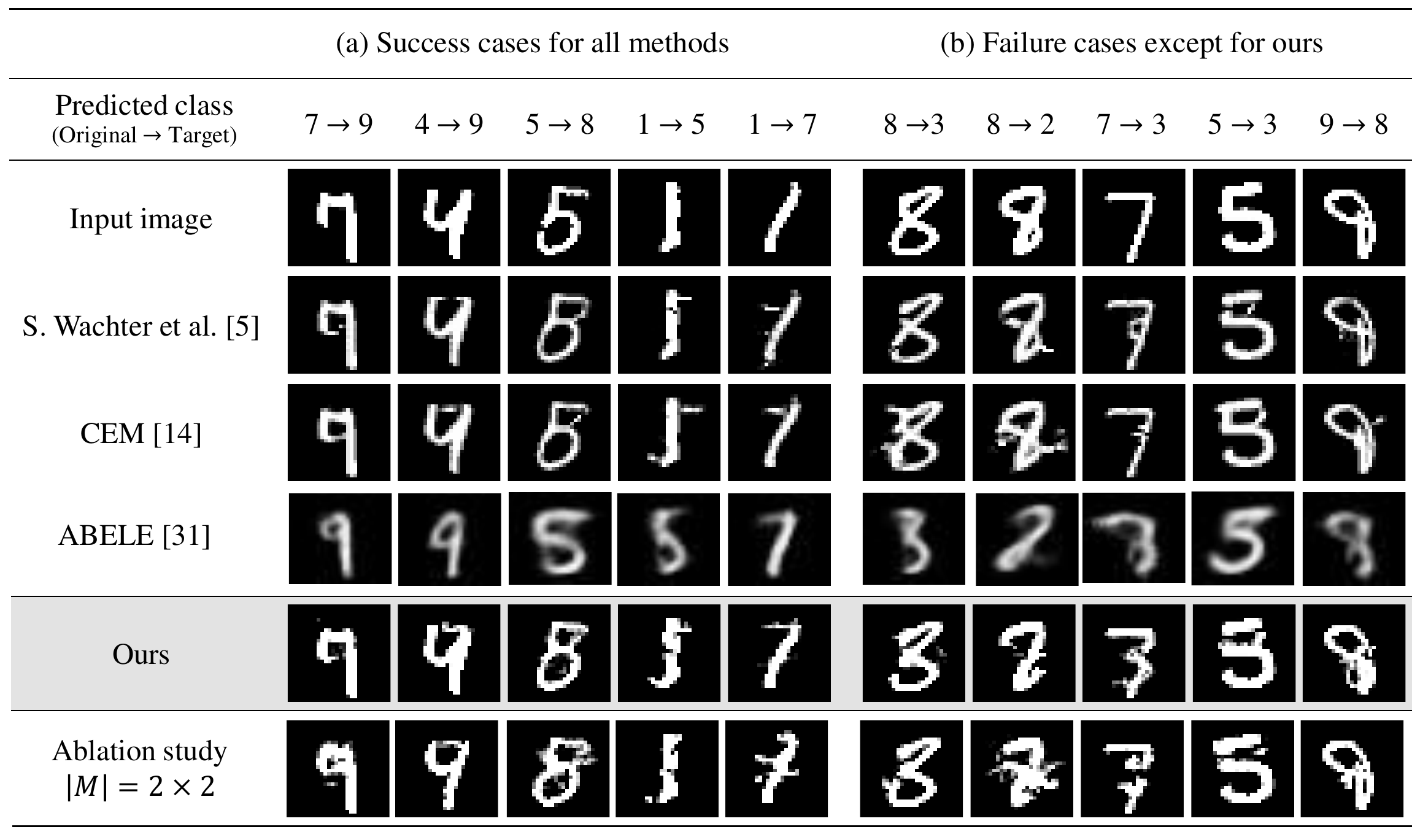}
   \caption{Counterfactual explanations on the image domain with the MNIST dataset \cite{lecun1998gradient}. From the input images, they generate perturbed images to classify them into a target class. (a) All methods show the success cases of counterfactual explanations. (b) Other counterfactual methods fail to provide human-friendly explanations unlike the proposed method.}
	\label{fig:MNISTResult_Comp}
   \end{center}
\end{figure*}

\subsubsection{Algorithmic details}
We performed a block-wise optimization for counterfactual explanation, considering that the adjacent pixels within an image share redundant information. In particular, instead of finding one pixel for the masking step, we set the dimension of the mask $M$ to $4 \times 4$ and we optimize the $16$ pixels in the composition step. Fig. \ref{fig:Making_MNIST} depicts the details of the process. Furthermore, to generate a visually smoothed image during the generation process, we added the total-variation regularization in Eq.~\ref{eq:ProposedLoss} similar to that in \cite{mahendran2015understanding}.
\begin{equation} \label{eq:Proposed_Loss_in_Image}
\argmin_{C} \left\Vert \sum_{k=1}^{K} f'_{k}(X')-\frac{1}{N}\sum_{i=1}^{N}  f'_{k}(\bar{X}_{i,c_t}) \right\Vert_2+\lambda \left\Vert X' - X \right\Vert_2 +\eta R_{tv},
\end{equation}
\noindent where $R_{tv}=\sum_{i,j} \left( \left( X^{'}_{i,j+1}-X^{'}_{i,j}\right)^{\beta} +\left( X^{'}_{i+1,j}-X^{'}_{i,j}\right)^{\beta} \right)^{\frac { \beta}{ 2}}$ and $\eta$ is a hyperparameter. The $i$ and $j$ are the indexes of the height and the width in an image. We set $\eta$ and $\beta$ to 0.3 and 2, respectively. The target probability $\tau$ is 0.9.

\subsubsection{Results}
For the comparison of our approach and existing counterfactual explanation methods; S. Wachter et al. \cite{wachter2018counterfactual}, CEM \cite{dhurandhar2018explanations} and ABELE \cite{guidotti2019black}, we randomly select test images and then analyze their explainability. Also, we set the target class as a similar class (e.g., 7 $\rightarrow$ 9, 5 $\rightarrow$ 8, 1 $\rightarrow$ 7) and a more difficult class to be changed (e.g., 1 $\rightarrow$ 5, 8 $\rightarrow$ 2, 7 $\rightarrow$ 3). Fig.~\ref{fig:MNISTResult_Comp}(a) shows that all methods successfully generate counterfactual explanations. They produce similar perturbed images for each original image and the important regions for a target class are well identified in terms of a human’s point of view. However, in Fig.~\ref{fig:MNISTResult_Comp}(b), S. Wachter et al. \cite{wachter2018counterfactual} and CEM \cite{dhurandhar2018explanations} produce adversarial images, so that it is difficult to interpret the results and identify the discriminative regions between the two classes. Meanwhile, ABELE \cite{guidotti2019black} produces blurred images. Conversely, our method exactly visualizes what features should be inserted and/or removed to be classified as the target class. It is worth noting that $|M|= 2 \times 2$ produces worse counterfactual images than $|M| = 4 \times 4$ as it increases the difficulty of the optimization by the redundancy among adjacent pixels.

\begin{figure*}[!t]
\begin{center}
\includegraphics[width=.5\linewidth]{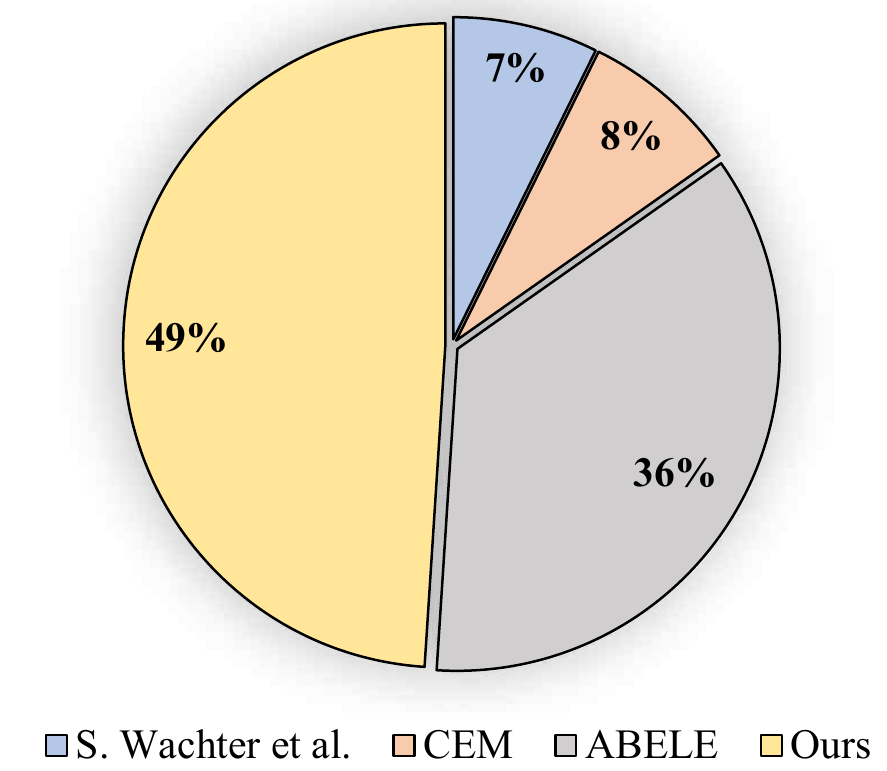}
   \caption{Evaluating how well human-friendly explanation is generated. Conducting an online survey, the statistic was obtained by calculating the percentage of counterfactual images identified as each method among the total images selected from 30 subjects. Details can be found in the text.}
	\label{fig:human_evaluation}
   \end{center}
\end{figure*}

In addition, to evaluate how well human-friendly explanation is generated, an online survey was configured as follows. First, 25 images were randomly selected for each algorithm to generate counterfactual images. After that, since this paper considers four algorithms, 100 counterfactual images, which were mixed randomly, were asked to check with 30 subjects whether those images looked like the same number as the ground-truth label. Finally, the statistic was obtained by calculating the percentage of counterfactual images identified as each method among the total images selected from 30 subjects. As shown in Fig. \ref{fig:human_evaluation}, the proposed method achieves the highest number as human evaluation.

\begin{figure}[t]
\begin{center}
		\includegraphics[width=0.8\linewidth]{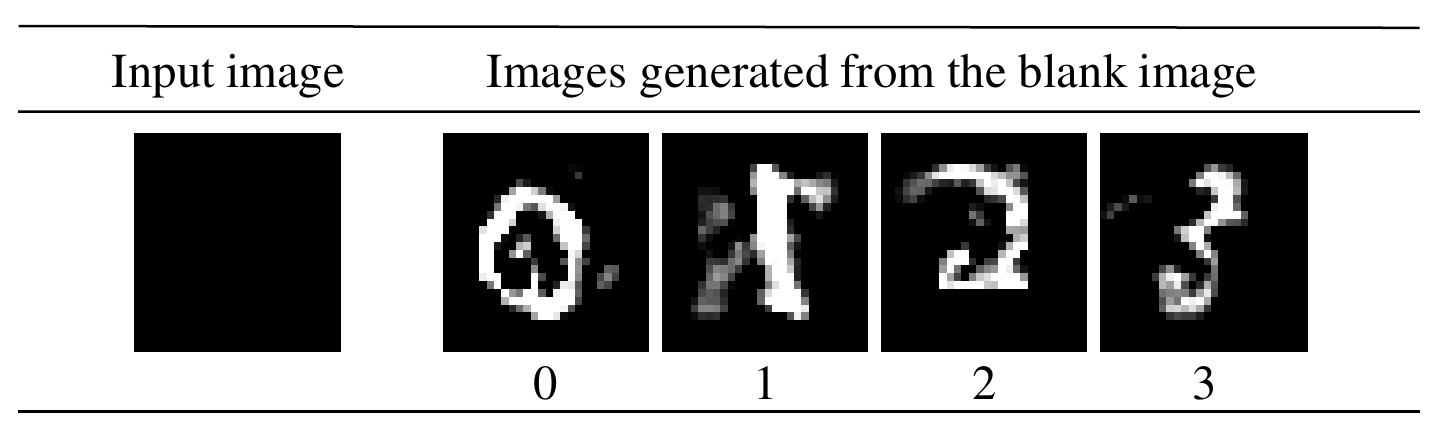}
\caption{Generation capability of the proposed method. From the black image, our method can even generate perturbed images that seem to be target classes by using Eq.~\ref{eq:Proposed_Loss_in_Image} except for the L2 regularization term. The target classes are presented by the numbers below the perturbed images.}
\label{fig:MNISTResult_from_BlackImage}
\end{center}
\end{figure}

To further verify the capability of feature generation, we provide experimental results by setting $\lambda$ of Eq.~\ref{eq:Proposed_Loss_in_Image} to zero in the composition step. As shown in Fig.~\ref{fig:MNISTResult_from_BlackImage}, our method even converts the black image into the perturbed images that seem to be target classes. That is, rendering the logit score of a perturbed data to belong to the logit space of training images that are classified as a target class can lead to generating similar images to the training data.

\subsection{HELOC and UCI Credit Card}
Both HELOC (Home Equity Line of Credit) \cite{FICO} and UCI Credit Card \cite{yeh2009comparisons} are tabular datasets for loan approval/refusal classification. Normalized input features are newly generated to change the classification result to the alternative class for counterfactual explanation. 

\subsubsection{Pre-trained network}
We normalized the input values into a range of [0, 1] using the training datasets and trained an MLP model, which is composed of five fully-connected layers. Test accuracies for the HELOC and UCI Credit Card datasets were 70.8\% and 81.0\%, respectively.  

\begin{table}[!htbp]
\caption{Counterfactual explanations on the tabular domain with the HELOC dataset \cite{FICO}. The changed feature values are represented in green.} 
\label{Tabular_Qualitative_HELOC}
\centering
\begin{adjustbox}{width=1\textwidth}

\begin{tabular}{c c ccc}
\hline
\footnotesize{\textbf{Feature name}} & \footnotesize{\textbf{Input data}}&\scriptsize{\textbf{S. Wachter et al. \cite{wachter2018counterfactual}}}& \footnotesize{\textbf{CEM \cite{dhurandhar2018explanations}}}& \footnotesize{\textbf{Ours}} \\[0.8ex]  \hline
    \scriptsize{MSinceOldestTradeOpenv} &  \small{0.427}    &  \small{ \textbf{\textcolor{OliveGreen}{0.426}}} & \small{0.427} &  \small{0.427} \\[0.8ex]

    \scriptsize{MSinceMostRecentTradeOpen} & \small{0.057}& \small{\textbf{\textcolor{OliveGreen}{0.052}}} & \small{0.057}&\small{0.057}\\[0.8ex]
    \scriptsize{AverageMInFile }& \small{0.465}&\small{\textbf{\textcolor{OliveGreen}{0.516}}}&\small{\textbf{\textcolor{OliveGreen}{0.504}}}&\small{0.465}                  \\[0.8ex]
    \scriptsize{NumSatisfactoryTrades } & \small{0.418}& \small{0.418}&\small{0.418}&\small{0.418} \\[0.8ex]
    \scriptsize{NumTrades60Ever2DerogPubRec } & \small{0.071}&\small{0.071}&\small{0.071}&\small{0.071} \\[0.8ex]
    \scriptsize{NumTrades90Ever2DerogPubRec } &    \small{0}  & \small{0}&\small{0}&\small{0} \\[0.8ex]
    \scriptsize{PercentTradesNeverDelq } & \small{0.88} & \small{0.88}&\small{0.88}&\small{0.88}  \\[0.8ex]
    \scriptsize{MSinceMostRecentDelq } & \small{0.348} & \small{0.348}&\small{0.348}&\small{0.348}  \\[0.8ex]
    \scriptsize{MaxDelq2PublicRecLast12M } & \small{0.667}& \small{\textbf{\textcolor{OliveGreen}{0.670}}}&\small{0.667}&\small{0.667}  \\[0.8ex] 
    \scriptsize{MaxDelqEver}  & \small{0.5} &\small{\textbf{\textcolor{OliveGreen}{0.498}}}&\small{0.5}&\small{0.5} \\[0.8ex] 
    \scriptsize{NumTotalTrades } & \small{0.398} & \small{0.398}&\small{0.398}&\small{0.398} \\[0.8ex] 
    \scriptsize{NumTradesOpeninLast12M  }& \small{0.125}&\small{0.125}&\small{0.125}&\small{0.125}\\[0.8ex] 
    \scriptsize{PercentInstallTrades } &  \small{0.55} & \small{0.55}&\small{\textbf{\textcolor{OliveGreen}{0.544}}}&\small{0.55} \\[0.8ex] 
    \scriptsize{MSinceMostRecentInqexcl7days } & \small{0.25} & \small{0.25}&\small{0.25}&\small{0.25}\\[0.8ex] 
    \scriptsize{NumInqLast6M } & \small{0}&\small{0}&\small{0}&\small{\textbf{\textcolor{OliveGreen}{0.057}}}\\[0.8ex] 
    \scriptsize{NumInqLast6Mexcl7days}  & \small{0}& \small{0}&\small{\textbf{\textcolor{OliveGreen}{-0.03}}}&\small{0} \\[0.8ex] 
    \scriptsize{NetFractionRevolvingBurden}  & \small{0.507}& \small{\textbf{\textcolor{OliveGreen}{0.498
    }}}&\small{0.507}&\small{0.507}\\[0.8ex] 
    \scriptsize{NetFractionInstallBurden}  & \small{0.491}&\small{\textbf{\textcolor{OliveGreen}{0.492}}}&\small{0.491}&\small{0.491}\\[0.8ex] 
    \scriptsize{NumRevolvingTradesWBalance  }& \small{0.3}&  \small{0.3}&\small{\textbf{\textcolor{OliveGreen}{0.298}}}&\small{0.3} \\[0.8ex] 
    \scriptsize{NumInstallTradesWBalance } &    \small{0.619} &  \small{\textbf{\textcolor{OliveGreen}{0.615}}}& \small{0.619}& \small{0.619}\\[0.8ex] 
    \tiny{NumBank2NatlTradesWHighUtilization}  & \small{0.476} & \small{0.476} &\small{0.476} &\small{0.476}  \\[0.8ex] 
    \scriptsize{PercentTradesWBalance } &  \small{0.787}& \small{0.787}&\small{0.787}&\small{0.787}\\[0.8ex] \hline
    \textcolor{black}{Prediction}  &  \small{\textcolor{red}{Not loanable}} &  \small{\textcolor{blue}{Loanable}}   &\small{\textcolor{blue}{Loanable}}     & \small{\textcolor{blue}{Loanable}} \\[0.8ex] \hline
\end{tabular}
\end{adjustbox}
\end{table}

\begin{table}[!htbp]
\caption{Counterfactual explanations on the tabular domain with the UCI Credit Card dataset \cite{yeh2009comparisons}. The changed feature values are represented in green.} 
\label{Tabular_Qualitative_UCI}
\centering

\begin{adjustbox}{width=0.9\textwidth}
\small
\begin{tabular}{ccccc}
\hline
 \footnotesize{\textbf{Feature name}} & \footnotesize{\textbf{Input data}}&\scriptsize{\textbf{S. Wachter et al. \cite{wachter2018counterfactual}}}& \footnotesize{\textbf{CEM \cite{dhurandhar2018explanations}}}& \footnotesize{\textbf{Ours}} \\  \hline
    \scriptsize{LIMIT BAL} &  \small{0.113}    &  \small{\textbf{\textcolor{OliveGreen}{0.111}}} & \small{0.113} &  \small{0.113} \vspace{-0.1cm}\\ 
    \scriptsize{SEX} & \small{1.0}& \small{1.0} & \small{1.0}&\small{1.0}\vspace{-0.1cm}\\ 
    \scriptsize{EDUCATION}&\small{0.166}&\small{\textbf{\textcolor{OliveGreen}{0.168}}}&\small{0.166}&\small{0.166}                  \\ 
    \scriptsize{MARRIAGE } & \small{0.333}& \small{0.333}&\small{0.333}&\small{0.333} \\ 
    \scriptsize{AGE } & \small{0.148}&\small{0.148}&\small{0.148}&\small{0.148} \\ 
    \scriptsize{PAY 0 } &    \small{0.4}  & \small{\textbf{\textcolor{OliveGreen}{0.479}}}&\small{\textbf{\textcolor{OliveGreen}{0.452}}}&\small{\textbf{\textcolor{OliveGreen}{0.423}}} \\ 
    \scriptsize{PAY 2} & \small{0.444} & \small{0.444}&\small{\textbf{\textcolor{OliveGreen}{0.446}}}&\small{0.444}  \\ 
    \scriptsize{PAY 3 } & \small{0.0} & \small{0.0}&\small{0.0}&\small{0.0}  \\ 
    \scriptsize{PAY 4  } & \small{0.0}& \small{0.0}&\small{0.0}&\small{0.0}  \\ 
    \scriptsize{PAY 5}  & \small{0.0} &\small{0.0}&\small{0.0}&\small{0.0} \\ 
    \scriptsize{PAY 6} & \small{0.0} & \small{0.0}&\small{0.0}&\small{0.0} \\
    \scriptsize{BILL AMT1 }& \small{0.026}&\small{0.026}&\small{0.026}&\small{0.026}\\ 
    \scriptsize{BILL AMT2} &  \small{0.042} & \small{0.042}&\small{0.042}&\small{0.042} \\ 
    \scriptsize{BILL AMT3 } & \small{0.076} & \small{\textbf{\textcolor{OliveGreen}{0.074}}}&\small{0.076}&\small{0.076}\\ 
    \scriptsize{BILL AMT4 } & \small{0.075}&\small{\textbf{\textcolor{OliveGreen}{0.073}}}&\small{0.075}&\small{0.075}\\ 
    \scriptsize{BILL AMT5}  & \small{0.132}& \small{0.132}&\small{0.132}&\small{0.132} \\ 
    \scriptsize{BILL AMT6}  & \small{0.397}& \small{0.397}&\small{0.397}&\small{0.397}\\ 
    \scriptsize{PAY AMT1}  & \small{0.0}&\small{0.0}&\small{\textbf{\textcolor{OliveGreen}{0.002}}}&\small{0.0}\\ 
    \scriptsize{PAY AMT2  }& \small{0.0}&  \small{0.0}&\small{\textbf{\textcolor{OliveGreen}{-0.012}}}&\small{0.0} \\ 
    \scriptsize{PAY AMT3} &    \small{0.0} &  \small{\textbf{\textcolor{OliveGreen}{0.004}}}& \small{0.0}& \small{0.0}\\ 
    \scriptsize{PAY AMT4}  & \small{0.0} & \small{0.0} &\small{0.0} &\small{0.0}  \\ 
    \scriptsize{PAY AMT5 } &  \small{0.0}& \small{0.0}&\small{\textbf{\textcolor{OliveGreen}{-0.004}}}&\small{0.0}\\ 
    \scriptsize{PAY AMT6 } &  \small{0.0}& \small{0.0}&\small{0.0}&\small{0.0}\\ \hline 
    \textcolor{black}{Prediction}  &  \small{\textcolor{red}{Not loanable}} &  \small{\textcolor{blue}{Loanable}}   &\small{\textcolor{blue}{Loanable}}     & \small{\textcolor{blue}{Loanable}} \\ \hline
\end{tabular}
\end{adjustbox}
\end{table}

\begin{table}[!t]
\centering
\caption{Quantitative evaluation using the L1, L2 and coherence metrics on the HELOC \cite{FICO} and UCI Credit Card \cite{yeh2009comparisons} datasets. The numbers indicate the mean and standard deviation.}
\begin{adjustbox}{width=1\textwidth}
\renewcommand{\arraystretch}{1.1}
\label{table_QuantitativeResult}
\Huge
\begin{tabular}{c|cc|cc|cc}
\hline
\multirow{2}[2]{*}{\textbf{Method}} & \multicolumn{2}{c|}{\textbf{L1 metric}} & \multicolumn{2}{c|}{\textbf{L2 metric}} & \multicolumn{2}{c}{\textbf{Coherence}} \bigstrut[t]\\
  & HELOC & UCI Credit Card & HELOC & UCI Credit Card & HELOC & UCI Credit Card \bigstrut[b]\\
\hline
S. Wachter et al. \cite{wachter2018counterfactual} & 16.25$\pm$4.54 & 4.55$\pm$5.54 & 4.52$\pm$0.21 & 0.17$\pm$0.09 & 2.34$\pm$2.66 & 2.06$\pm$2.48 \bigstrut[t]\\
CEM \cite{dhurandhar2018explanations} & 4.12$\pm$1.89 & 3.12$\pm$2.36 & 0.27$\pm$0.17 & 0.20$\pm$0.11 & 1.27$\pm$0.36 & 1.94$\pm$1.96 \\
Growing Sphere \cite{laugel2018comparison} & 9.37$\pm$2.60 & 6.83 $\pm$ 2.37 & 0.18$\pm$0.13 & 0.12 $\pm$ 0.09 & 1.10$\pm$0.19 & 1.48 $\pm$ 1.17 \\
LORE \cite{guidotti2019factual} & 1.19 $\pm$ 0.95 & 1.35 $\pm$ 1.18 & 0.68 $\pm$ 1.04 & 0.37 $\pm$ 0.39 & 6.97 $\pm$ 3.21 & 6.48 $\pm$ 13.14 \bigstrut[b]\\
\hline
Ours & 1.05 $\pm$ 0.10 & 1.01 $\pm$ 0.12 & 0.39 $\pm$ 0.19 & 0.33 $\pm$ 0.17 & 1.40 $\pm$ 0.17 & 1.29 $\pm$ 0.47 \bigstrut\\
\hline
\end{tabular}%
\end{adjustbox}
\end{table}

\subsubsection{Algorithmic details}
For these tabular datasets, Algorithm \ref{pseudocode} is used as is, that is, without undergoing further processes for explanation. The target probability $\tau$ was set to 0.5.

\subsubsection{Results}
Tables \ref{Tabular_Qualitative_HELOC} and \ref{Tabular_Qualitative_UCI} detail the qualitative examples and prove that our method can effectively change the decision of the pre-trained network with fewer feature modifications compared to other methods. These results can be useful when our counterfactual explanations are applied commercially in financial institutions. For example, suppose a customer who hopes to obtain a loan from a bank, but the AI system of the bank refused to grant the loan based on the record of the customer. In this situation, the customer may ask how to achieve loan approval. Fortunately, our counterfactual method can provide the important factors (e.g., loan amount and credit score) for the decision and how the values of the features should be modified for the loan approval. Furthermore, as the proposed method perturbs fewer input features than existing methods, the loan can be granted to the customer by changing only a small amount of information.

To further provide quantitative results, we introduce two L1 and L2 metrics to measure the minimality property. The L1 metric aims to count the number of perturbed features from the original data as 
\begin{equation}
\phi_{1}=\mathbbm{1}_{[0.001,\infty]}(X_{o,i}-X'_{o,i}), 
\end{equation}
where $X_{o,i}$ is the $i$-th element of the original data, $\mathbbm{1}$ is an indicator function and the lower bound $0.001$ is used to ignore noise values produced by the generative process.

The L2 metric measures the difference of a value between the perturbed data and the original data as
\begin{equation}
\phi_{2}=||X_{o,i}-X'_{o,i}||_{2}.
\end{equation}

In addition, a robustness metric is further considered for performance comparison. Specifically, for the tabular datasets of the financial domain, it is crucial to provide two users that have similar personal information with similar counterfactual features. Thus, we added the Lipswhic estimation as
\begin{equation}
    robustness(X_o) = \argmax_{X_i \in N(X_o)} \dfrac{ \left\Vert X'_i - X'_o \right\Vert_2 }{  \left\Vert X_i - X_o \right\Vert _2},
\end{equation}
where $N(X_o) = \{ X_i \in \mathbbm{X} \, | \, \sum_{j} \mathbbm{1}_{[0.001,\infty]}\left(|X_{i,j} - X_{o,j}|\right) \leq \epsilon \}$, $X_i$ is the instances in the test set $\mathbbm{X}$ and $X_{i,j}$ is the $j$-th element of $X_i$. We call this setting coherence by following R. Guidotti \textit{et al.} \cite{guidotti2019black}.

We evaluate the three metrics on $1,000$ randomly selected test samples and the experimental results are shown in Table~\ref{table_QuantitativeResult}. As compared to other methods, we can observe that our method generally not only uses fewer features to change the original decision on both datasets but also achieves low coherence values. In other words, our method generates similar counterfactual explanations for two analogous user information while using only a few input features.


\section{Ablation study}
\subsection{Effect of logit distributions}
To demonstrate that considering the logit distributions of the training data is crucial in generating counterfactual explanations, we compare the results using our loss function in Eq. \ref{eq:ProposedLoss} with the following loss function without exploiting the logit score distribution. 
\begin{equation}\label{eq:PriorLoss_Ablation}
\argmax_{C} f_{c_{t}}((1-M)\circ X+M\circ C)-\lambda \left\Vert X' - X \right\Vert_2.
\end{equation}
Thus, Eq. \ref{eq:PriorLoss_Ablation} principally aims to increase the classification probability for $c_t$.

\begin{table}[t]
\begin{adjustbox}{width=1\textwidth}
\begin{tabular}{ccccc}
\hline
               & \textbf{Text data} &  \textbf{Prediction} & \textbf{Text data} &  \textbf{Prediction} \\ \hline 
Input text &  \makecell{It could be one of the best \\ movies of the year}
        &  \textcolor{red}{Positive} & \makecell{There are many other funny \\ scenes in this film} & \textcolor{red}{Positive}\\ 

Ours w/o LD &  \makecell{It could be one of the \textbf{\textcolor{blue}{Feroz}} \\ movies of the year}
        &  \textcolor{blue}{Negative} & \makecell{There are many other \textbf{\textcolor{blue}{Equally}} \\ scenes in this film} & \textcolor{blue}{Negative} \\

Ours &  \makecell{It could be one of the \textbf{\textcolor{blue}{mediocre}} \\ movies of the year}
        &  \textcolor{blue}{Negative} & \makecell{There are many other \textbf{\textcolor{blue}{bad}} \\ scenes in this film} & \textcolor{blue}{Negative} \\ \hline
\end{tabular}
\end{adjustbox}
\caption{Ablation study on randomly generated text data. We used a network trained on the IMDB dataset \cite{maas2011learning}. The perturbed word from an original word is represented in blue. Our method without considering the logit distribution (Ours w/o LD) produces the word that is irrelevant to the target class. However, by considering the logit distribution (Ours), the perturbed text is highly related to that of the target class.
}
\label{fig:Ablation_IMDB}
\end{table}

\begin{figure}[t!]
\begin{center}
		\includegraphics[width=0.8\linewidth,height=0.3\linewidth]{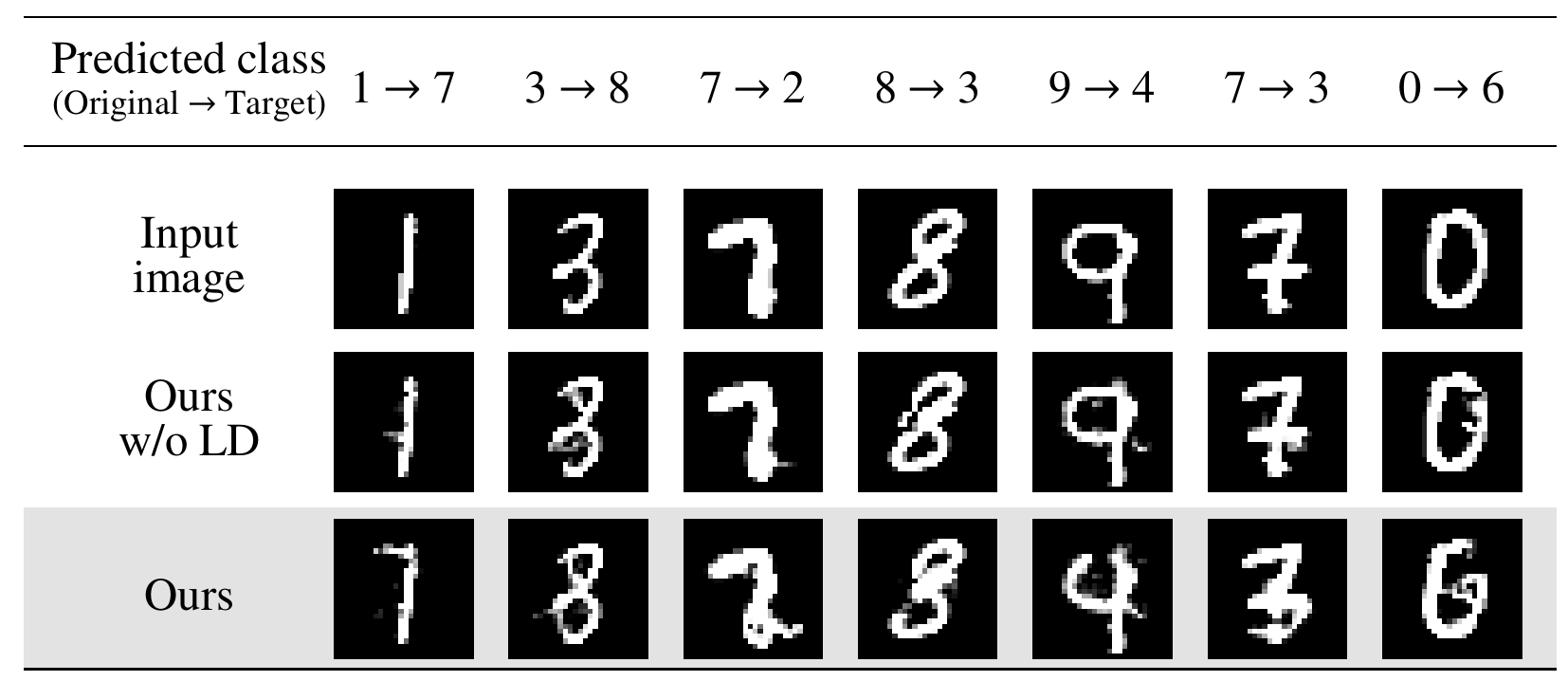}
\caption{Ablation study on the MNIST dataset \cite{lecun1998gradient}. Ours w/o LD denotes the method that does not consider the logit distribution and it generates the images that are classified as the target classes but seem to be adversarial data. On the other hands, our method produces more human-acceptable images for the target classes, thus providing what features are discriminative between two classes.}
\label{fig:Ablation_MNIST}
\end{center}
\end{figure}

The results for the IMDB sentiment analysis dataset \cite{maas2011learning} and the MNIST dataset \cite{lecun1998gradient} are presented in Table \ref{fig:Ablation_IMDB} and Fig. \ref{fig:Ablation_MNIST}, respectively. As shown in Table \ref{fig:Ablation_IMDB} (left), the method not using the logit distribution generates a perturbed text data ``Feroz'' that is an actor's name to be classified as the negative class. On the other hand, we can observe that our method changes the word in the original text into the pertinent and interpretable word to be classified as the target class. Likewise, for the MNIST dataset, the method without considering the logit distribution makes the results look like adversarial data, so that we cannot interpret which regions are discriminative between the original and target classes. Meanwhile, our method generates the results that seem to be the digit images for the target classes. To summarize, we show that the proposed loss function can prevent counterfactual explanation from being adversarial data, and generates more human-friendly interpretation for the characteristics of a pre-trained model.

\subsection{Applying a factual explanation method}
\begin{figure*}[!t]
\begin{center}
\includegraphics[width=.7\linewidth]{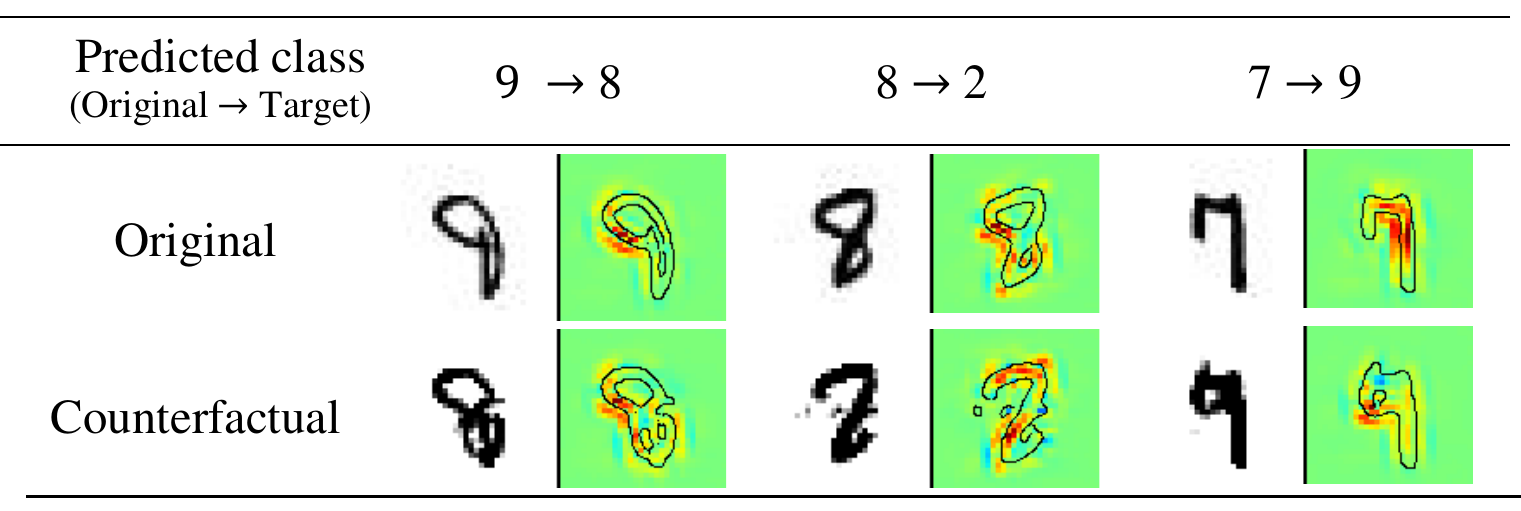}
   \caption{{Applying LRP \cite{bach2015pixel} to counterfactual explanation. We can observe that an factual explanation model can be combined with counterfactual images to figure out feature importance.}}
	\label{fig:cf-lrp}
   \end{center}
\end{figure*}

Counterfactual features of each class can be combined with factual explanation methods such as LRP \cite{bach2015pixel} to figure out feature importance. In other words, a saliancy map for the counterfactual explanation of $8 \rightarrow 2$ should indicate important pixels to be classified as the target class $2$. To this end, we show that the generated counterfactual images can be combined with external factual explanation methods in Fig. \ref{fig:cf-lrp}.

\section{Discussion}
As we only provide quantitative comparisons on tabular datasets, we discuss the reasons as follows. Unlike factual explanation such as LRP \cite{bach2015pixel} that does not alter input features but highlights the most important parts, counterfactual explanation should \textit{generate} input features that change the classification result. This causes several issues for each dataset. For MNIST, as shown in the right part of Fig. 5, there exist several cases where an image, which looks like an original class but is classified as another class, is generated. In other words, we have the risk of adversarial attacks and measuring quantitative metrics such as the minimality property for non-interpretable images are not possible to provide useful information. This is clearly different characteristics from factual explanation. For IMDB, let's see a toy example. If a counterfactual method changes ``the person is \textbf{awful}'' to ``the person is \textbf{good}'' and another is to ``the person is \textbf{nice}'', it is unnatural to compare the distances between `\textbf{awful} -- \textbf{good}' and `\textbf{awful} -- \textbf{nice}'.  For HELOC and UCI Credit Card, unlike the MNIST and IMDB datasets, the perturbed input features themselves contain important information. For example, if a counterfactual method suggests that the salary of a person should be changed from \$2,000 to a \$2,550, the difference of \$550 itself is crucial for loan approval. Thus, the quality of an algorithm can be quantitatively measured by the number of changed input features and the degree of changes.

\section{Limitation and future work}
Although the proposed counterfactual method is model-agnostic and can handle multi-class classification, there are a few methods that can provide both factual and counterfactual explanations simultaneously, such as LORE \cite{guidotti2019factual} for a binary classifier and ABELE \cite{guidotti2019black} for the image domain. Those methods exploit a decision tree that is itself interpretble to obtain a local decision boundary that is approximately fitted around the input. However, due to the non-linearity that makes up the feature space, \textit{two similar inputs} can produce  different local decision boundaries, resulting in \textit{two different decision trees} and thus lower coherence values as shown in Table \ref{table_QuantitativeResult}. Nonetheless, providing both factual and counterfactual explanation is highly helpful to people in various domains. As an example, for a person who has been rejected from loan, an AI system can provide two explanations: ``Your loan request has been denied as \textbf{Salary $\leq \$1,000$} and \textbf{Age $>$ 50} but it can be approved if \textbf{Salary $> \$2,000$}''. In this view, our method only provides the latter. Thus, it is meaningful to combine the advantages of LORE \cite{guidotti2019factual} that can explain both factual and counterfactual factors and the proposed method that stably produces the L1 and coherence performances. We would like to leave this as future work.

\section{Conclusion}
In this paper, we proposed a counterfactual explanation based on gradual construction, which iterates over masking and composition steps. The masking step selects an important subset of features to classify a given input into a target class by using the directional derivative with respect to the original data. Then, the composition step updates the values of the selected features to not only improve the classification probability for a target class but also push the perturbed data towards being similar to input features. We showed that it is crucial to consider the logit distribution of training data for the composition step to prevent a perturbed data from being adversarial data. Experimental results also verified that our method satisfies explainability and minimality properties as it qualitatively provides more acceptable interpretation than the existing counterfactual explanation methods from a human's point of view and quantitatively uses fewer features for data generation.

\section*{Acknowledgment}
This work was supported by Institute for Information \& communications Technology Planning \& Evaluation(IITP) grant funded by the Korea government(MSIT) (No. 2017-0-01779, A machine learning and statistical inference framework for explainable artificial intelligence, and No. 2019-0-00079, Artificial Intelligence Graduate School Program, Korea University).

\bibliography{counterfactual}
\end{document}